\documentclass[twocolumn,a4paper]{article}
\usepackage[colorlinks, urlcolor=blue,
	pdffitwindow, bookmarksopen,
	filecolor=blue,
	bookmarksopenlevel=0,
	pdfpagelayout=SinglePage, driverfallback=dvipdfm]{hyperref}
\usepackage{ncsp23} 
\usepackage{color}
\usepackage{cite}
\usepackage{amsmath,amsfonts}
\usepackage{algorithmic}
\usepackage{multirow}
\usepackage{latexsym}
\usepackage{newtxtext,newtxmath}
\usepackage{algorithm}
\usepackage{enumitem}
\usepackage{here}
\usepackage{cite}
\usepackage{graphicx}
\usepackage{textcomp}
\usepackage{xcolor}
\usepackage{subcaption}
\usepackage{caption}
\usepackage{float}

\begin{document}

\title{A Privacy Preserving Method with a Random Orthogonal Matrix for ConvMixer Models.}

\name{
 \begin{tabular}{c}
Rei Aso$^1$ , Tatsuya Chuman$^1$ and Hitoshi kiya$^1$
 \end{tabular}
 }

\address{
\begin{tabular}{c}
$^1$Tokyo Metropolitan University\\
6-6 Asahigaoka, Hino-shi, Tokyo 191-0065, Japan\\
\end{tabular}
}




\maketitle

\section*{Abstract}

In this paper, a privacy preserving image classification method is proposed under the use of ConvMixer models. To protect the visual information of test images, a test image is divided into blocks, and then every block is encrypted by using a random orthogonal matrix. Moreover, a ConvMixer model trained with plain images is transformed by the random orthogonal matrix used for encrypting test images, on the basis of the embedding structure of ConvMixer. The proposed method allows us not only to use the same classification accuracy as that of ConvMixer models without considering privacy protection but to also enhance robustness against various attacks compared to conventional privacy-preserving learning.

\section{Introduction}

Deep learning has been deployed in many applications including security-critical ones. Generally, data contains sensitive information such as personal informational, so privacy-preserving methods for deep learning have become an urgent problem [1]. To achieve privacy-preserving learning, various methods have been proposed. One of them is Federated Learning (FL) [2], which is a type of distributed learning. FL allows us to train a model over multiple participants without directly sharing their raw data. However, FL have not considered the protection of test data in cloud environments so far. In this paper, we propose a novel method for protecting visual information on test images.

 To protect visual information on plain images in untrusted cloud environments, many learnable encryption methods have been studied so far [3]-[13]. Learnable encryption has to satisfy three requirements in general: (a) having a high accuracy that is almost the same as that of plain models, (b) being robust enough against various attacks, and (c) easily updating a secret key. However, most of existing methods [3]-[11] degrade the accuracy of models due to the use of encrypted images, and moreover, need to retrain models to update the key. 
In contrast, the similarity between block-wise encryption and the architecture of isotropic networks has been pointed out to enable us to perfectly stratify the two requirements that the existing methods cannot [12][13]. Information on embeddings in isotropic networks such as the vision transformer [14] and ConvMixer [15] is encrypted by random matrixes generated with secret keys for privacy-preserving learning. However, in the conventional methods [12][13], simple permutation matrixes are used for image and model encryption, so encrypted images are not robust enough against various attacks. Accordingly, we propose the use of a novel random matrix, which is called a random orthogonal one generated by using the Gram-Schmidt orthonormalization. The proposed method allows us to enhance the visual protection of images, while maintaining the same as that of plain models and the easy update of a secret key.   

\section{ConvMixer}
Before discussing the proposed method, we summarize ConvMixer and its properties briefly.
ConvMixer is mainly used for image classification tasks and is known for its high classification performance[15].
The structure of ConvMixer is inspired by the Vision Transformer (ViT)[14].
ViT consists of two Embedding processes (Patch Embedding and Position Embedding) and a Transformer structure.
On the other hand, ConvMixer consists of a Patch Embedding and a CNN structure.
Figure 2 shows the structure of ConvMixer, which consists of two main structures: Patch Embedding and ConvMixer Layer.
In this paper, we focus on Patch Embedding.
In Patch Embedding, an input image $x \in \mathbb{R}^{H \times W \times C} $ of height $H$, width $W$, and number of channels $C$ is divided into patches of size $p \times p$.
Each patch is then transformed into a vector $x_p^i \in \mathbb{R}^{p^2C}$, multiplied by the learnable filter $E$
and linearly transform it into a vector of $d$-dimensions by taking the product of $x_p^i \in \mathbb{R}^{p^2C}$ as
\begin{gather}
  z = [x_p^{1}E,...,x_p^{i}E,...,x_p^NE] \\
 z \in \mathbb{R}^{N \times d} , E \in \mathbb{R}^{(p^2c) \times d} \nonumber
  \end{gather}

In previous studies[12][13], it is known that it is possible to protect the privacy of test images by transforming the filter $E$ with a secret key.
In this paper, we propose a method to achieve stronger privacy preserving of test images by using random orthogonal matrices.
\begin{figure}[ht]
    \centering
    \vspace{-100pt}
    \includegraphics[bb=45 0 1004 550,scale=0.35]
    {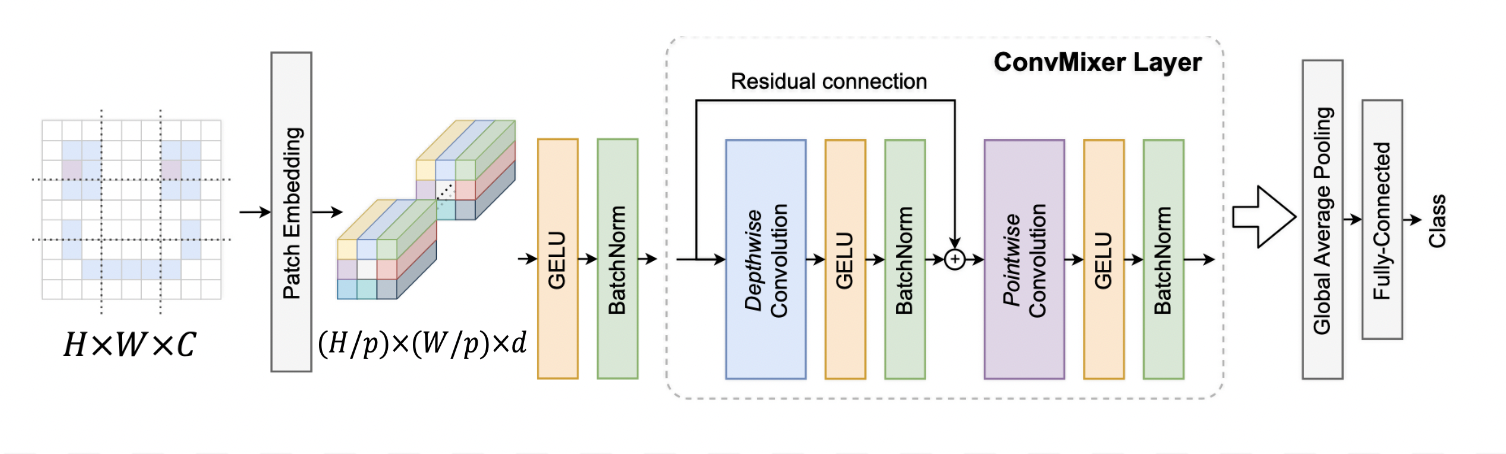}
    \caption{Architecture of ConvMixer}
\end{figure}
 
\section{Proposed Method}
\subsection{Overview}

\begin{figure}[ht]
    \centering
    \includegraphics[bb=0 0 1004 550,scale=0.35]{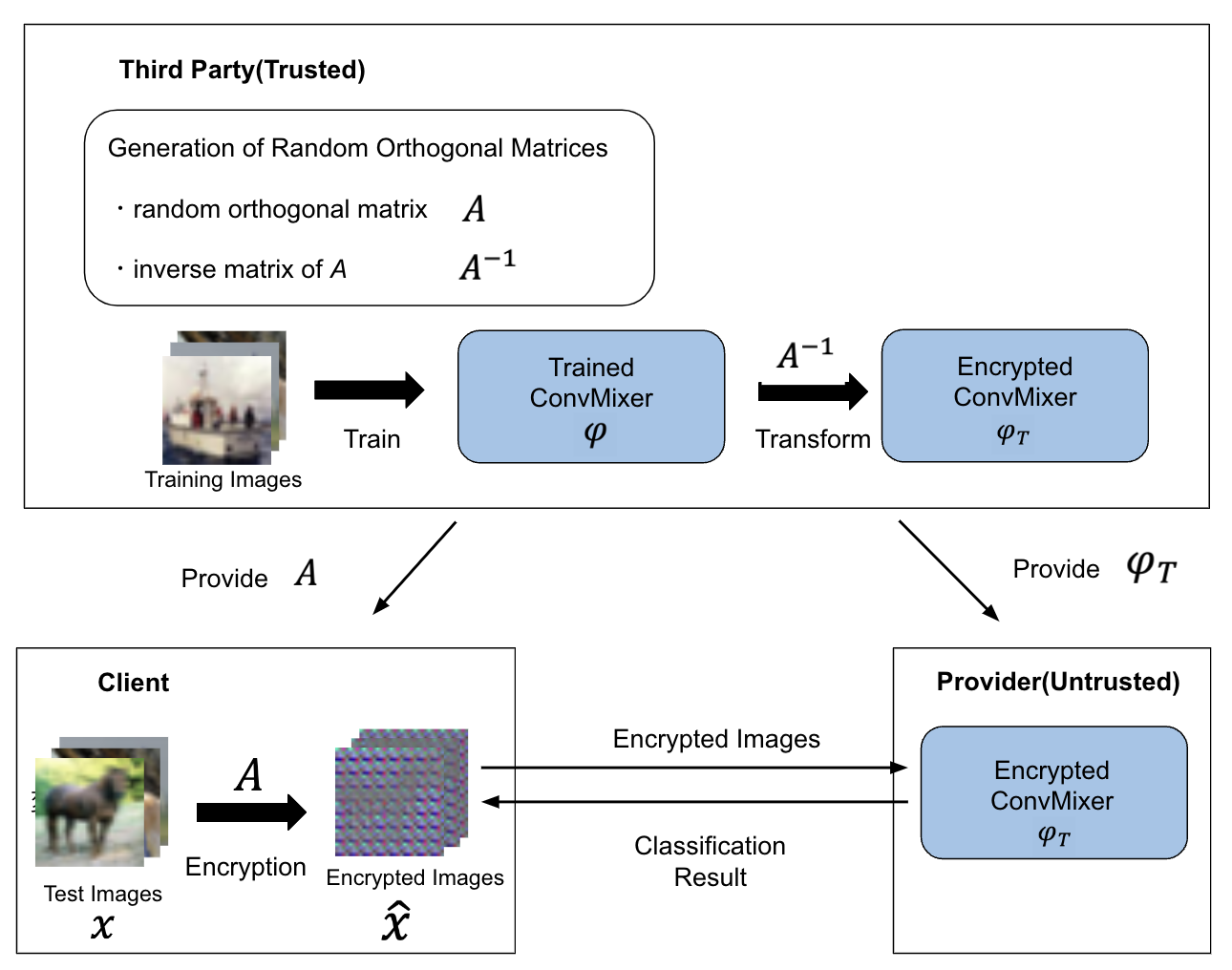}
    \caption{Framework of proposed method}
\end{figure}

Figure 2 illustrates the framework of the proposed method. The proposed method aims to protect visual information on test images. To achieve this aim, we encrypt test images and a transform model by using an random orthogonal matrix.
The framework is summarized as below.
\begin{quote}
 \begin{itemize}
  \item A third party (trusted) generates random numbers with a secret key (seed), and prepares a random orthogonal matrix $A$ from the random numbers and an inverse random orthogonal matrix $A^{-1}$.
  \item The third party trains a ConvMixer model $\varphi$ with plane images. The trained model $\varphi$ is transformed into an encrypted model $\varphi_{T}$ by using $A^{-1}$.
  \item The third party provides the random orthogonal matrix $A$ to a client (trusted) and model $\varphi_{T}$ to a provider (untrusted).
  \item The client transforms a test image $x$ into an encrypted image $\hat{x}$ by using $A$. After that, the client sends $\hat{x}$ to the provider.
  \item The provider inputs $\hat{x}$ into model $\varphi_{T}$, and sends back a prediction result to the client.
 \end{itemize}
\end{quote}
Even if the provider is not trusted, the client does not give visual information of test images and matrix $A$ used for image encryption to the provider. Thus, the client can receive prediction results while maintaining the privacy preserving of test images.
\subsection{Test Image Encryption}
A test image $x \in \mathbb{R}^{H \times W \times C} $ is transformed into an encrypted image $\hat{x} \in \mathbb{R}^{H \times W \times C} $ as below.
\begin{enumerate}
  \setlength{\parskip}{0cm} 
  \setlength{\itemsep}{0.1cm} 
  \item Divide $x$ into $N$ blocks with a size of $p \times p$ such that $B = \{B_1,...,B_N \}$, where $p \times p$ is the same size as the patch size used in a ConvMixer model, and $N$ is $(H \times W)/p^2$.
  \item Flatten each block $B_i \in \mathbb{R}^{p\times p \times C}$ into a vector $x_p^i \in \mathbb{R}^{p^2C} $ as 
   \begin{gather}
     x_p^i = [x_p^i(1),..., x_p^i(p^2C) ].
   \end{gather}
  \item Generate a encrypted vector $\hat{x}_p^i \in \mathbb{R}^{p^2C} $ by multiplying vector $x_p^i$ by matrix $A \in \mathbb{R}^{(p^2 C) \times {(p^2 C)}} $ as 
   \begin{gather}
     \hat{x}_p^i = x_p^i A.
   \end{gather}
  \item Rebuild vector $\hat{x}_p^i$ into block $\hat{B}_i$ in the reverse order of step 2.
  \item Concatenate $\hat{B}=\{\hat{B}_1,...,\hat{B}_N \}$ into an encrypted test image $\hat{x}$.
\end{enumerate} 
\subsection{Model Encryption}
To avoid the performance degradation caused by encryption of test images, $E$ in Eq.(1) is transformed  by using $A^{-1}$ as
\begin{gather}
   E' = A^{-1}E .
\end{gather}
When replacing $E$ and $x_p^i$ with $E'$ and $\hat{x}_p^i$ , respectively,
vector z in Eq.(1) is reduced to as
\begin{gather}
    z' = [\hat{x}_p^{1}E',...,\hat{x}_p^{i}E',...,\hat{x}_p^NE'] .
\end{gather}
Thus, by substituting Eqs. (3) and (4) to Eq.(5), we obtain:
\begin{gather}
    z' = [x_p^{1}AA^{-1}E,...,x_p^{i}AA^{-1}E,...,x_p^NAA^{-1}E] \nonumber \\
    = [x_p^{1}E,...,x_p^{i}E,...,x_p^{N}E] = z .
\end{gather}
From Eq.(5), encrypted model $\varphi_{T}$ allows us to have the same performance as that of the model trained with plane images, under the use of encrypted images.
\subsection{Generation of Random Orthogonal Matrices}
A random orthogonal matrix $A$ can be generated by using the Gram–Schmidt orthonormalization.
The procedure for generating $A$ with a size of $n \times n$ is given as follows.
\begin{enumerate}
   \item[1.]Generate an real matrix $R$ with a size of $n \times n$ by using a random number generator with a seed.
   \item[2.]Calculate $det(R)$, and proceed to 3 if $det(R)\neq 0$. Otherwise, return to 1.
   \item[3.]Compute a random orthogonal matrix $A$ from $R$ by using the Gram-Schmidt orthogonalization. 
\end{enumerate}
In this framework, any regular matrix can be used as A for image encryption. Several conventional methods for privacy-preserving image classification use permutation matrices of pixel values, in which many elements have zero values in matrices as 
   \begin{gather}
    A = 
    \begin{bmatrix}
        0 & 1 & 0\\
        0 & 0 & 1\\
        1 & 0 & 0\\
    \end{bmatrix}
.
   \end{gather} In contrast, the proposed random orthogonal matrices include no the zero values as elements. The use of such matrices allows us not only to more strongly protect visual information on plain images but to also enhance robustness against various attacks, while maintaining the same performance as that of models trained with plain images. In addition, $A^{-1}$ can easily be calculated as the transposed matrix of A.
\section{Experiment Results}
To verify the effectiveness of the proposed method, we ran a number of experiments on the CIFAR-10 dataset. 
\subsection{Setup}
We used the CIFAR-10 dataset, which consists of 60,000 color images (dimension of 32 × 32 × 3) with 10 classes (6000 images for each class) where 50,000 images are for training and 10,000 for testing. ConvMixer was trained and tested on the CIFAR-10 dataset. In the model setting, we set the patch size to 4, the number of channels after patch embedding to 256, the kernel size of depth-wise convolution to 7 and the number of ConvMixer layers to 8. models were trained for 200 epochs with the Adam optimizer, where the learning rate was 0.001. We also used a random orthogonal matrix with a size of 48 × 48 for the encryption of test images and models.

\subsection{Visual Protection Performance}
Figure 3 shows an example of images encrypted with a conventional encryption method[12][13], in which pixel shuffling and negative-positive transformation are carried out for image encryption, and an example of images encrypted with the proposed method, where the images had $H \times W \times C=512 \times 512 \times 3$ as an image size, and the block sizes used for encryption were $p=8$ and $p=16$. When using an orthogonal matrix for encryption, transformed pixel values are real values, so (b) in Fig. 3 were displayed after normalizing the pixel values to the range of [0.1]. From the figures, the selection of a larger the block size gave smaller visual information. The use of random orthogonal matrices was also demonstrated to have a stronger visual protection performance than that of the conventional method. 
In addition to visual protection, encrypted images have to be robust enough against various attacks, which aim to restore visual information from encrypted images. We already confirmed that images encrypted with the proposed method are more robust against attacks including 
jigsaw puzzle solver attacks [16]. In particular, unlike ViT, ConvMixer models do have position embedding, so the position of patches cannot be changed. Therefore, privacy-preserving ConvMixer needs a stronger encryption method than ViT.

\begin{figure}[ht]
\hspace{-15pt}
\scalebox{0.75}[0.75]{
    \begin{tabular*}{50mm}{@{\extracolsep{\fill}}c|c|c}
        \begin{minipage}{4truecm}
             \centering
              \includegraphics[bb=-30 280 280 280,scale=0.275]{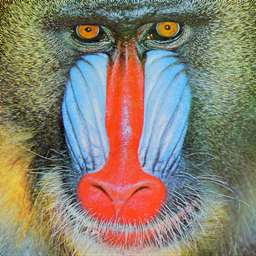}
            \end{minipage}
        &
        \begin{minipage}{4truecm}
             \centering
              \includegraphics[bb=0 0 550 550,scale=0.125]{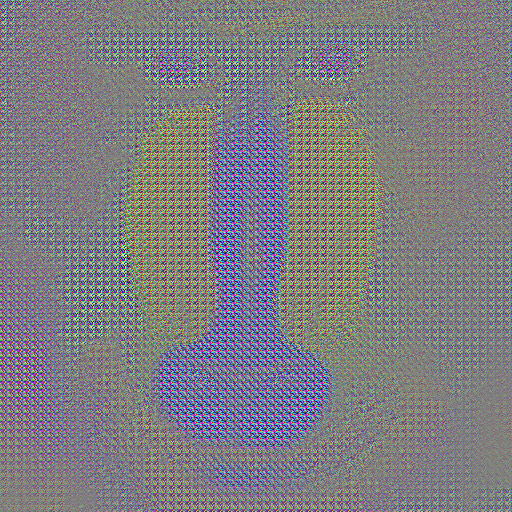}\\
              \subcaption{\footnotesize{p = 8}}
            \end{minipage}
        &
        \begin{minipage}{4truecm}
             \centering
              \includegraphics[bb=0 0 550 550,scale=0.125]{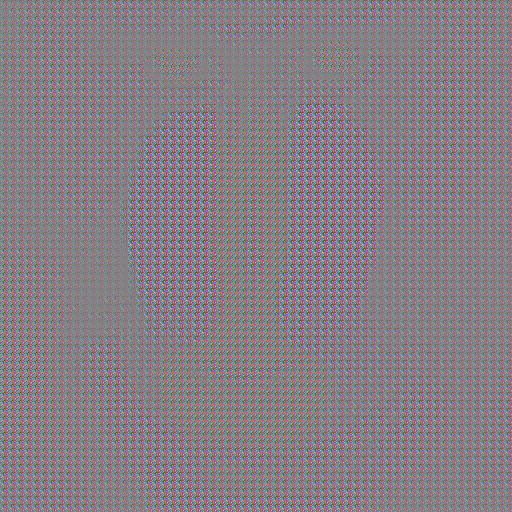}\\
              \subcaption{\footnotesize{p = 8}}
            \end{minipage}\\
            
        &
        \begin{minipage}{4truecm}
             \centering
              \includegraphics[bb=0 0 550 550,scale=0.125]{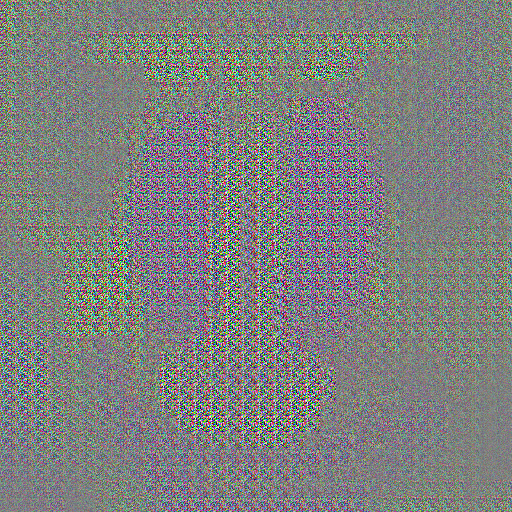}\\
              \subcaption{\footnotesize{p = 16}}
              \vspace{10pt}
            \end{minipage}
        &
        \begin{minipage}{4truecm}
             \centering
              \includegraphics[bb=0 0 550 550,scale=0.125]{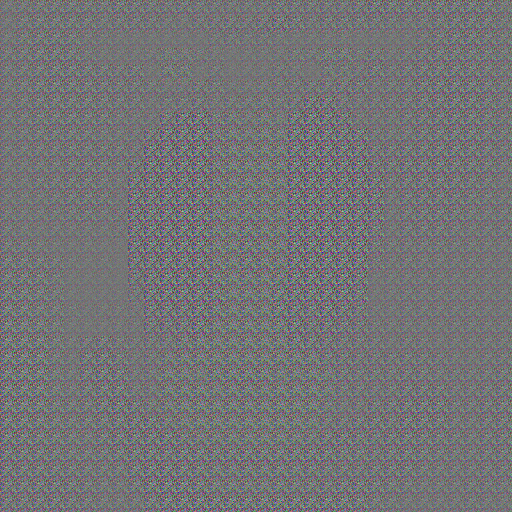}\\
              \subcaption{\footnotesize{p = 16}}
              \vspace{10pt}
            \end{minipage}\\
        Plane image& (a) & (b) \\
    \end{tabular*}
    }
    \caption{Example of encrypted images with (a) conventional method[12][13] and (b) proposed method}
\end{figure}

\subsection{Classification Performance}
We evaluated the classification performance of the proposed method as shown in Table 1, where plain and encrypted indicate plane test images and an encrypted test images, respectively, and plain model and encrypted model are models trained with plain images, and models trained with encrypted images.  
Table 1 shows the classification results for each combination. From the table, the proposed method (the combined use of Encrypted models and encrypted images) had the same classification accuracy as that of the baseline without privacy protection (plain models and plain images). Accordingly, the proposed method can not only protect the visual information of test image, but also classify the encrypted image without any degradation of classification accuracy.
\vspace{-5pt}
\begin{table}[ht]
\caption{Classification accuracy (\%)}
\label{table:SHF}
\centering
\begin{tabular}{cccll}
\cline{1-3}
\multicolumn{1}{c|}{model\textbackslash{}test image} & plane                & encypted           &  &  \\ \cline{1-3}
\multicolumn{1}{c|}{plane model}                     & 90.38                & 13.2                 &  &  \\
\multicolumn{1}{c|}{encypted model}                      & 10.27                & 90.38                &  &  \\ \cline{1-3}
\multicolumn{1}{l}{}                                 & \multicolumn{1}{l}{} & \multicolumn{1}{l}{} &  & 
\end{tabular}
\end{table}
\\
\section{Conclusion}
In this paper, we proposed a novel method for protecting visual informational on test images under the use of ConvMixer models. The proposed method allows us to use a random orthogonal matrix for image encryption, and it was demonstrated not only to enhance the visual protection of images but to also maintain the same accuracy as that of models trained with plain images.

\section*{Ackowledgment}
This study was partially supported by JSPS KAKENHI (Grant Number JP21H01327).

\end{document}